\documentclass[runningheads]{llncs}
\usepackage[T1]{fontenc}
\usepackage{url}
\usepackage[usenames,dvipsnames,table]{xcolor}
\usepackage{amsmath}
\usepackage{amssymb}
\usepackage{mathtools}
\usepackage{graphicx}
\usepackage[colorlinks=true, allcolors=blue]{hyperref}
\usepackage{algpseudocode}
\usepackage{algorithm}
\usepackage{setspace}
\usepackage{booktabs}
\usepackage{subcaption}
\usepackage{tabularray}
\usepackage{hyperref}
\usepackage{tikz}
\usepackage{multirow}
\usetikzlibrary{bayesnet, calc, arrows.meta, bending}
\usetikzlibrary{decorations.pathreplacing,angles,quotes,shapes.multipart, calligraphy}
\usepackage{amsfonts}
\usepackage{array}
\usepackage{hhline}
\usepackage{url}
\usepackage{longtable}

\usepackage{soul}

\usepackage[colorlinks=true]{hyperref}
\begin{document}
%Memory-Efficient
\title{Latent Causal Modeling for 3D Brain MRI Counterfactuals}
% \thanks{Supported by organization x.}
%
\titlerunning{Latent Causal Modeling for 3D Brain MRI Counterfactuals}
% \author{Anonymized Authors}  %% Added for anonymized MICCAI 2025 submission
% \authorrunning{Anonymized Author et al.}
% \institute{Anonymized Affiliations \\
%     \email{email@anonymized.com}}
\author{Wei Peng\inst{1} \and Tian Xia\inst{3} \and Fabio De Sousa Ribeiro\inst{3} \and Tomas Bosschieter\inst{1} \\
Ehsan Adeli\inst{1} \and
Qingyu Zhao\inst{2} \and
Ben Glocker\inst{3} \and  
Kilian M. Pohl\inst{1}}
% index{Peng, Wei \and Adeli,Ehsan \and Bosschieter, Tomas  \and Park, Sang Hyun  \and Zhao, Qingyu  \and Pohl, Kilian M.}
% Department of Psychiatry and Behavioral Sciences, Stanford University, Stanford, United States of America
\authorrunning{ Peng et al.}
% % % First names are abbreviated in the running head.
% % % If there are more than two authors, 'et al.' is used.
% % % % 

\institute{{\small Stanford University, Stanford, CA 94305} \and Weill Cornell Medicine, New York, NY 10065 \and
 Imperial College London, South Kensington Campus
London SW7 2AZ, UK
}
% \affil[1]{}
% \affil[2]{{\small Department of Computer Science, Stanford University, Stanford, CA 94305}}
% \affil[3]{{\small Center for Biomedical Sciences, SRI International, Menlo Park, CA 94205}}

\maketitle              
% typeset the header of the contribution \it{e.g.,}
%
\begin{abstract}
The number of samples in structural brain MRI studies is often too small to properly train deep learning models. Generative models show promise in addressing this issue by effectively learning the data distribution and generating high-fidelity MRI. However, they struggle to produce diverse, high-quality data outside the distribution defined by the training data. One way to address the issue is using causal models developed for 3D volume counterfactuals. However, accurately modeling causality in high-dimensional spaces is a challenge so that these models generally generate 3D brain MRIS of lower quality. To address these challenges, we propose a two-stage method that constructs a Structural Causal Model (SCM) within the latent space. In the first stage, we employ a VQ-VAE to learn a compact embedding of the MRI volume. Subsequently, we integrate our causal model into this latent space and execute a three-step counterfactual procedure using a closed-form Generalized Linear Model (GLM). Our experiments conducted on real-world high-resolution MRI data (1mm) provided by the Alzheimer's Disease Neuroimaging Initiative (ADNI) and the National Consortium on Alcohol and Neurodevelopment in Adolescence (NCANDA) demonstrate that our method can generate high-quality 3D MRI counterfactuals.

% and look more realistic than the ones produced by GAN-based models. 
% data set with long-tail distribution
%\keywords{Generative Model  \and 3D MRI generation \and Neuroimaging.}
\end{abstract}

\section{Introduction}
Generative AI models have demonstrated great potential to advance numerous medical fields including neuroimaging with 3D brain MRIs~\cite{peng2023generating,pinaya2022brain}. 
The synthesis of medical images is particularly promising for tasks like improving image quality, imputing missing modalities \cite{zheng2022diffusion}, and modeling disease progression~\cite{zhao2021longitudinal,jung2021conditional,jung2023conditional}.
Recent advances in diffusion probabilistic models (DPMs)~\cite{ho2020denoising,sohl2015deep} have played a significant role in this development as they are able to accurately capture the underlying data distributions. This results in synthesized medical images containing a great amount of detail~\cite{wolleb2022diffusion,la2022anatomically} that also differ from the training samples~\cite{karras2019style,xing2021cycle}. However, a key challenge remains to generate realistic samples that are outside the distribution defined by the training data set~\cite{kwon2019generation}. Overcoming this challenge is essential to the generalizability of large-scale models~\cite{xing2021cycle}. 
% Furthermore, existing models are incapable of answering causal questions like ``\textit{what would a brain have looked like had its factor $X$ been $x$ instead of $y$?}" in a principled manner. 
One way to address this shortcoming is by incorporating causality between factors during the generation progress~\cite{pearl2009causality}. For example, aging causes a thinning of the cortex that is accelerated in those with alcohol use disorder or Alzheimer's disease, which provides a causal interaction not taken into account by statistical models. Here, we propose to integrate causality in the synthesis of 3D brain MRIs so that their generation can be guided by (interventions on) factors such as age, brain regions, and diagnosis.

Specifically we rely on probabilistic Structural Causal Model (SCM)~\cite{pearl2009causality,peters2017elements}, which can incorporate the relationships between variables, directed from cause to effect. It indicates that intervening on a cause should lead to changes in the effect, rather than the other way around. This forces the model to consider our premises and assumptions about causality~\cite{fabio2023cfmodel}.
% To leverage these probabilistic SCMs for counterfactual generation of MRIs specifically, we propose a novel model 
%
% In our case, $\mathbf{x}\in X$ is high-dimensional MRIs, which are caused by lower dimensional endogenous parent variables $\mathbf{pa}_\mathbf{x} \subseteq X \setminus \{\mathbf{x}\}$ (e.g. factors like age). The set of mechanisms in $F$ are learned using deep learning components inspired by DSCMs~\cite{pawlowski2020deep} and NCMs~\cite{xia2021causal,xia2023neural}. 
Contrary to previous SCMs~\cite{pawlowski2020deep,xia2023neural}, we focus on the synthesis of 3D high-fidelity image counterfactuals of structural t1w MRIs. However, building an SCM in a high-dimensional space presents considerable computational challenges, particularly when dealing with voluminous 3D MRIs. To that end, we build a counterfactual model that can learn the causality and perform counterfactual generation in a latent space, which allows us to produce 3D brain MRIs of higher quality than other counterfactual models.

% To address this issue, we propose constructing a counterfactual brain generation model~\cite{fabio2023cfmodel} that  We do so as cohorts defined according to this metadata can greatly differ with respect to the shape (including volume) of brain regions. For example, aging causes a thinning of the cortex that is accelerated in those with alcohol use disorder or Alzheimer's disease. Our method combines the causality and deep representation learning~\cite{Bengio2013,scholkopf2021toward,pawlowski2020deep,xia2023neural}. 

Our model first employs a VQ-VAE \cite{vqvae2017} to encode the high-dimensional 3D brain MRI into a low-dimensional latent space. We then integrate a causal graphical model~\cite{fabio2023cfmodel,pawlowski2020deepscm} into this latent space such that interventions are performed in the latent space rather than the observation space (which contains MRIs). The three-step counterfactual procedure, i.e., abduction, action, and prediction, is subsequently achieved by a Generalized linear Model (GLM)~\cite{lu2021metadata} with a closed-form solution.

To the best of our knowledge, this is the first work that can perform high-fidelity counterfactual generation of 3D T1w MRIs. Our counterfactual generative model does not only improve interpretability and generative performance but can also diversify (MRI) datasets. Additionally, counterfactual explanations could be key for preventive purposes, as it is inherently capable of showing how one's brain might change under certain `conditions' such as prolonged substance use. Our main contributions can be summarised as follows:
\begin{itemize}
     \item We present a novel causal generative modeling framework in a latent space for producing high-fidelity 3D MRI counterfactuals with Markovian probabilistic causal models;
    \item Our model fulfills all three rungs of Pearl’s ladder of causation and the three-step procedure of the counterfactual generation can be realized efficiently by a novel GLM method;
    \item We show that our model can perform high fidelity 3D counterfactual generation and also demonstrate its \textit{axiomatic soundness} of our counterfactuals by evaluating its context-level generation. 
\end{itemize}

\section{Methodology}\label{method}
\begin{figure}[!t]
    \centering
    \includegraphics[width=0.98\textwidth]{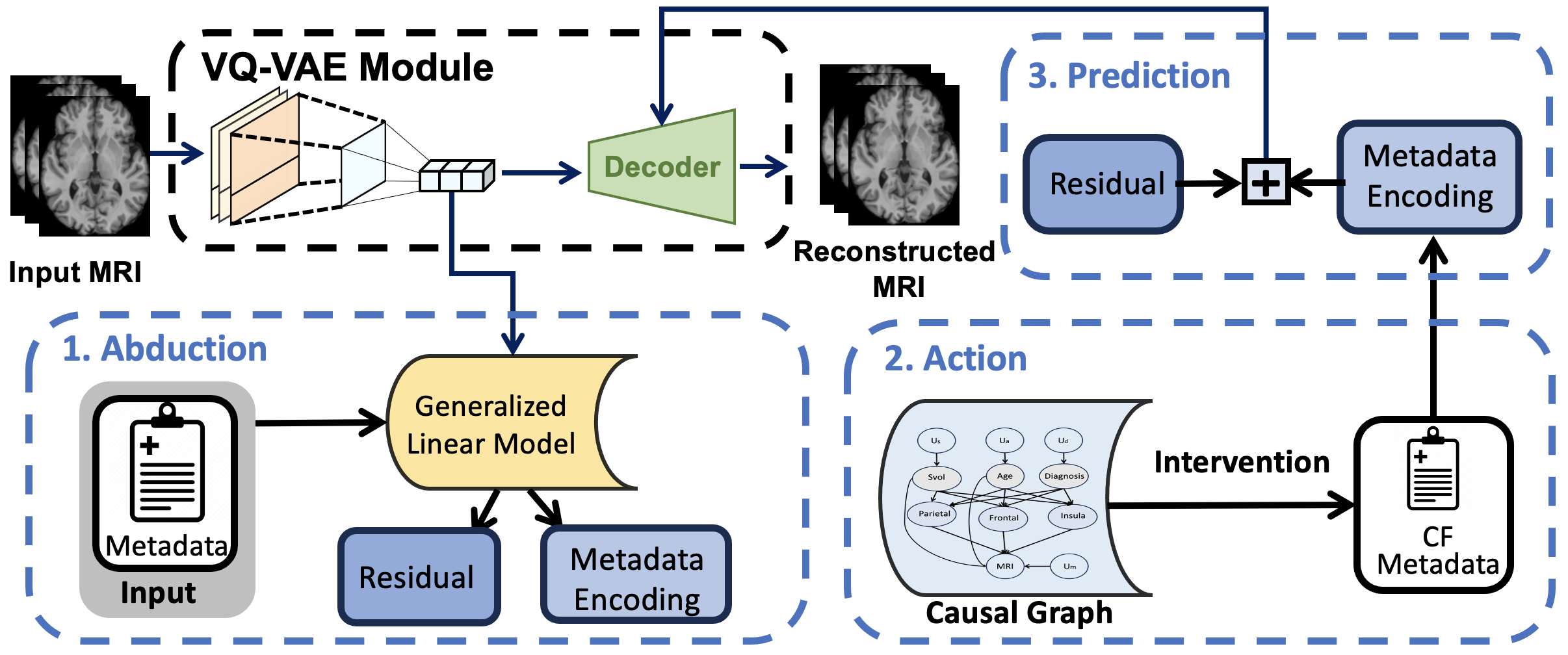}
    \caption{Architecture of the LSCM for image: Stage I (black box) consists of a VQ-VAE to encode the real 3D MRI to a quantized vector representation. Based on this, in stage II (blue boxes), a Latent SCM is constructed. The three-step procedure of counterfactual inference is achieved by a efficient GLM. }
    \label{fig:method}
\end{figure}

Our method comprises a two-stage process (outlined in Fig.~\ref{fig:method}) that performs counterfactual generation of MRIs given attributes, e.g., age and brain regions of interest (ROIs). Given an intervention on (at least) one of these attributes, a \textit{Deep Structural Causal Model} (DSCM) deploys a causal graph to compute the counterfactual of the attributes. For example, consider a brain MRI scan of an 80-year-old female, from which we extract the attributes \verb|age=80|, \verb|sex=female|, and the two ROI scores \verb|ROI_1_score=0.3| and \verb|ROI_2_score=0.7|. Now to generate the brain MRI at age 50 years old, we intervene and set \verb|age=50|, after which the DSCM computes the counterfactual attribute values, e.g., \verb|age=50|, \verb|sex=male|, \verb|ROI_1_score=0.4| and \verb|ROI_2_score=0.6|. These counterfactual attributes are then used to compute the counterfactual feature embeddings, which are fed into a VQ-VAE decoder \cite{vqvae2017} in order to generate the corresponding MRI. These two stages are discussed below. 
% OLD
% Our counterfactual model is a novel generative model (Fig.~\ref{fig:method}), which learns causal mechanisms in the latent space and  synthesizes high-resolution 3D MRI counterfactuals. The proposed model mainly contains two components. The first component produces counterfactuals for low-dimensional variables, i.e., attributes. The produced attribute counterfactuals are used to guide the generation of MRI counterfactuals in the second component, which introduces a Generalized Linear Model (GLM) into a VQ-VAE~\cite{vqvae2017}. Below we detail the two components.

%%% OLD OLD
% In the first stage, the MRI is turned into a quantized feature encoding, which is generated by Vector Quantization coupled with a Variational Autoencoder (VQ-VAE)~\cite{vqvae2017}. The second stage computes counterfactuals following a three-step procedure~\cite{pearl2009causality}, which is achieved by a probabilistic structural causal model (SCM) and a Generalized Linear Model~\cite{mcnamee2005regression} (GLM). Based on the output latent counterfactual, a brain MRI is further synthesized by feeding it to the generator of VQ-VAE. The individual components of our method are now reviewed in further detail.

\subsection{Deep Structural Causal Models}
\label{subsec:Deep Mechanisms for Structured Variables}
A structural causal model (SCM)~\cite{pearl2009causality} is a triple $\mathcal{M} \coloneqq \langle U, V, F \rangle$ consisting of two sets of variables: (i)  \textit{exogenous} variables $U = \{u_1,\dots,u_N\}$; (ii) \textit{endogenous} variables $V = \{v_1,\dots,v_{N}\}$, and a set of functions known as \textit{causal mechanisms} $F = \{f_1,\dots,f_{N}\}$, which determine the values of the endogenous variables: $v_{k}:=f_{k}(\mathbf{pa}_{k}, u_{k})$, where $\mathbf{pa}_{k}$ are the direct causes (\textit{parents}) of $v_{k}$. It is possible to perform an intervention on any $v_{k}$ by, e.g., setting it to a constant $do(v_{k} \coloneqq c)$ and thereby disconnect $v_{k}$ from its causal parents. SCMs enable the estimation of \textit{counterfactuals} (hypothetical scenarios) via a three-step process: (i) \textit{abduction}: infer the posterior distribution $P(U \mid X)$, which represents the current state of the world; (ii) \textit{action}: perform one or more interventions to obtain a modified model; (iii) \textit{prediction}: infer counterfactual values of the endogenous variables using the modified model and the posterior $P(U \mid X)$.

In this work, we use a deep structural causal model (DSCM)~\cite{pawlowski2020deep} to model the causal relationships between patient attributes such as age, brain regions of interest, and diagnosis. Following \cite{de2023high}, a DSCM uses a conditional normalizing flow~\cite{rezende2015variational} as the mechanism for each endogenous variable $v_{k}:=f_{k}({u}_{k};\mathbf{pa}_{k})$, to enable tractable and explicit abduction of the exogenous noise $u_{k}=f_{k}^{-1}(v_{k};\mathbf{pa}_{k})$. DSCMs can thereby tractably estimate counterfactuals given observed attributes and intervention(s). The counterfactual of any $v_k$ is given by\\ $\widehat{v}_{k} \coloneqq f_{k}(f_{k}^{-1}(v_{k};\mathbf{pa}_{k}); \mathbf{\widehat{pa}}_k)$, where $\mathbf{\widehat{pa}}_k$ are the counterfactual parents of $v_k$ resulting from an upstream intervention.
% , where $\widehat{u}_{k}$ is the abducted noise and $\mathbf{\widehat{pa}}_k \subseteq \{\widehat{v}_{k} \}^{N}_{k=1,k\neq i}$ are counterfactuals of $v_{k}$'s parents.

However, it has been shown that directly applying DSCMs~\cite{pawlowski2020deep} to 3D brain MRIs is impractical due to the complexity of constructing and optimizing invertible neural networks~\cite{tabak2010density,rezende2015variational}. Therefore, we propose a method that operates in a low-dimensional latent space. We first use a Vector Quantized Variational Autoencoder (VQ-VAE) to learn a compressed latent representation of the MRIs. Within this space, our approach then adapts the Latent Structural Causal Model (LSCM) to separate the MRI variables ($\mathbf{x}$) from other relevant variables ($V$, e.g., clinical or demographic data). We then utilize a Generalized Linear Model (GLM)~\cite{lu2021metadata} to perform the counterfactual steps, such as abduction and action. The GLM is well-suited for this task as it provides an interpretable, linear framework for modeling these causal interventions. After estimation of all counterfactual attributes, we obtain the counterfactuals of image latent features $\mathbf{z}$'s parents (see next section for details of $\mathbf{z}$), i.e. $\mathbf{\widehat{pa}_{z}}$, which will be used to generate $\widehat{\mathbf{z}}$. The resulting counterfactual feature encoding is converted into an MRI using the decoder of the VQ-VAE model.

 % After completing training, our method synthesizes a new MRI counterfactual by first generating its latent exogenous variables, which are the residual of the GLM. Then, an action is performed on the causal graph such that the parent variables after the intervention are produced according to the functional relationships in the graph. Finally, prediction consists of adding the residual to the metadata-specific encoding, which is derived from metadata values from the causal graph. 

\subsection{Latent SCM for 3D Counterfactual MRIs}

To fully leverage the latent SCM described in Section~\ref{subsec:Deep Mechanisms for Structured Variables}, a high-dimensional MRI $\mathbf{x}$ first has to be encoded to a corresponding latent feature $\mathbf{z}$, for which we use a VQ-VAE. Then, we obtain its counterfactual $\mathbf{\widehat{z}}$ based on $\mathbf{\widehat{pa}_{z}}$, which is finally decoded to the MRI counterfactual $\mathbf{\widehat{x}}$. We now discuss the process of obtaining these latent features $\mathbf{z}$ and estimating the counterfactuals $\mathbf{\widehat{z}}$, after which decoding is trivial given a trained VQ-VAE.

\paragraph{\textbf{Computing latent feature} $\mathbf{z}$.} 
To obtain a latent feature $\mathbf{z}$ from a 3D MRI $\mathbf{x}$, we employ a Variational Autoencoder (VQ-VAE)~\cite{vqvae2017}. VQ-VAEs comprise an encoder \(\mathcal{E}\) and decoder \(\mathcal{D}\), as well as a code book \(\mathcal{C}\) that performs vector quantisation (VQ). That is, code book $\mathcal{C}$ empowers mapping a 3D MRI $\mathbf{x}$ to a quantized encoding. 

The first step in the process of computing the latent feature $\mathbf{z}$ is using the encoder \(\mathcal{E}\) to map the 3D MRI $\mathbf{x} \in \mathbb{R}^{D \times H\times W}$ to its lower-dimensional feature representation $\mathbf{z} \coloneqq \mathcal{E}(\mathbf{x}) \in \mathbb{R}^{d \times h\times w \times n_D}$. Correspondingly, the decoder \(\mathcal{D}\) aims to reconstruct a 3D MRI \(\hat{\mathbf{x}}=\mathcal{D}(\mathbf{z})\) from a compressed embedding \(\mathbf{z} \). The training objective of the model is to minimize the reconstruction error. Once latent feature $\mathbf{z}$ is obtained, we apply vector quantization, which discretizes the continuous latent space using a set of embedding units \( \mathcal{C} \coloneqq \{(k, e(k)): k = 1, 2, \ldots, N_{\mathcal{C}} \mbox{ and  } e(k) \in \mathbb{R}^{n_D}\}\). This allows an MRI to be represented very efficiently by a set of indices (a.k.a. code) as defined with respect to the code book. Notably, this process also serves a regularization purpose to avoid overfitting the model during training. 

After the encoding and vector quantization steps are finished, the latent representation \(\mathbf{z}\) is described by \(d \times h\times w \) vectors of \(n_D\) dimensions each. This is achieved by searching the nearest neighbor of the embedding units in codebook \(\mathcal{C}\) for each feature vector in \(z_c\). In other words, each of the \(d \times h\times w \) vectors \(z_c \in \mathbb{R}^{n_D}\) from \(\mathbf{z}\) is paired with its closest representative \(e(k)\) from the codebook such that the corresponding code (a.k.a. index) is 
\( \mathcal{I}(z_c):= \text{arg min}_{k}(z_c-e(k)).\) Subsequently, \(e(\mathcal{I}(z_c))\) is the representative such that the quantization of the latent encoding may be formulated as \(Q(z_c) = [e(\mathcal{I}(z_c))]_{z \dashv\,  \mathbf{z}}\), where $[ \cdot ]_{z_c \dashv\, \mathbf{z}}$ denotes the vectors $z_c$ defined according to $\mathbf{z}$.
Building on this foundation of vector quantization, we propose a fine-grained quantization process that acquires multiple quantizations per vector, namely the vector's own quantization as well as the quantization of its residual. However, we first reshape the MRI's latent encoding $\mathbf{z}\in \mathbb{R}^{d \times h\times w \times n_D}$ to $\mathbb{R}^{2\times d \times h\times w \times \frac{n_D}{2}}$ so that $\mathbf{z}$ consists of twice as many vectors \(z_h \in \mathbb{R}^{\frac{n_D}{2}}\) with half the size; in other words, each $z_c \in \mathbb{R}^{n_D}$ gets split into two vectors $z_h \in \mathbb{R}^{\frac{n_D}{2}}$. For each of the vectors $z_h$, we determine the code \(\mathcal{I}(z_{h})\) as done above (with \(e(k) \in \mathbb{R}^{\frac{n_D}{2}}\) also being half in size), and then compute the code corresponding to the residual, i.e., \(\mathcal{I}(z_{h}- e(\mathcal{I}(z_{h})))\),
% , i.e., 
% \[ \mathcal{I}_R(z) := [\mathcal{I}(z), \mathcal{I}(z- e(\mathcal{I}(z)))].\]
so that the quantized encoding can be written as 
\begin{equation}\label{eq:qz}
    Q(z_{h}) \coloneqq e(\mathcal{I}(z_{h}))+ e(\mathcal{I}(z_{h}- e(\mathcal{I}(z_{h}))).
\end{equation}
% \[ Q(z) = e(\mathcal{I}(z))+ e(\mathcal{I}(z- e(\mathcal{I}(z))) ).\]
% The two representations are stacked 
For every $z_c$ comprised of two $z_h$, we stack the two quantized encodings $Q(z_h)$ to reconstruct a vector of the original size, thus obtain a quantized vector representation of $\mathbf{z}$, which is subsequently fed into the decoder.

\paragraph{\textbf{Estimate latent counterfactual $\mathbf{\widehat{z}}$ and $\mathbf{\widehat{x}}$}.} Given feature \(\mathbf{z} = Q(\mathcal{E}(\mathbf{x})\)), we will build a closed-form mechanism \(f\) for \(\mathbf{z}\), such that its exogenous variable is given by $\mathbf{u}_\mathbf{z} = f^{-1}_\theta(\mathbf{pa}_{\mathbf{z}}, \mathbf{z})$. We achieve this by using the GLM. Specifically, we first flatten the feature as a vector with dimension \(K = d \times h\times w \times n_D\). Then, a feature matrix \(\mathbf{Z} \in \mathbb{R}^{N \times K} \) is built for all the \(N\) samples in the dataset. At the same time, the \(\mathbf{pa}_{\mathbf{z}}\) for all training samples is accumulated to construct a matrix \(\mathcal{P} \in \mathbb{R}^{N \times m}\), where \(m\) is the number of variables in  \(\mathbf{pa}_{\mathbf{z}}\). Then, an ordinary least square estimator~\cite{lu2021metadata} is applied to solve the GLM's normal equations so that the linear parameters \(\mathcal{B}\) can be represented by the closed-form solution
\[ \mathcal{B} = (\mathcal{P}^T \mathcal{P})^{-1}\mathcal{P}^T \mathbf{Z}.\] 
Subsequently, the \textbf{Abduction} step of the counterfactual generation can be represented as the computation of the residual component \(\mathbf{U}_\mathbf{Z} \in \mathbb{R}^{N \times K}\) acting as the exogenous variable for features  \(\mathbf{Z} \), which is given by
\[ \mathbf{U}_\mathbf{Z} = \mathbf{Z} - \mathcal{P}\mathcal{B} = (\mathbf{I} - (\mathcal{P}^T \mathcal{P})^{-1}\mathcal{P}^T ) \mathbf{Z}.\]
Given that we construct $\mathcal{P}$ explicitly, we find that the solution for $\mathcal{B}$, and thus for $\mathbf{U}_\mathbf{Z}$, is of closed-form.
This is computationally feasible as we perform the GLM in the latent space, not in the high-dimensional MRI space. Our method is scalable and easily lends itself to cases containing many more samples. Like metadata normalization~\cite{lu2021metadata}, we can accumulate the \(\mathcal{B}\) with momentum.  

The \textbf{Action} step of the counterfactual generation can be realized by using the causal graph to produce counterfactual parents \(\widehat{\mathbf{pa}}_{\mathbf{z}}\). Once we have the \(\widehat{\mathbf{pa}}_{\mathbf{z}}\), the \textbf{Prediction} step of counterfactual generation can be achieved by adding factor-specific embeddings back to the exogenous variable as follows:
\[ \widehat{\mathbf{Z}} = \mathbf{U}_\mathbf{Z} + \widehat{\mathcal{P}}\mathcal{B}.\]

By recovering its spatial resolution and decoding using the generator, we get the MRI counterfactual in the observation/MRI space, i.e., \(\mathbf{x} = \mathcal{D}(\hat{\mathbf{z}}), \hat{\mathbf{z}} \in \widehat{\mathbf{Z}}\).

\begin{figure*}[t!]
\centering
\begin{subfigure}{0.78\textwidth}
\centering
    \includegraphics[width=\textwidth]{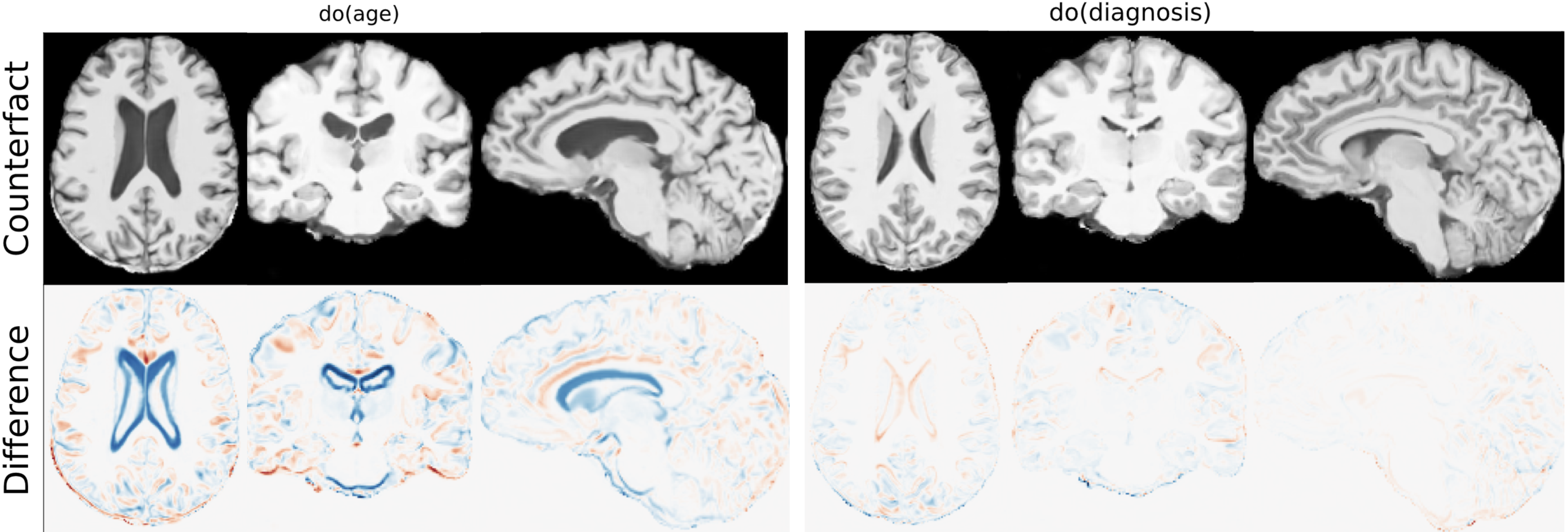}
\end{subfigure}
\hfill
\begin{subfigure}{0.16\textwidth}
        \centering
        \includegraphics[width=\textwidth]{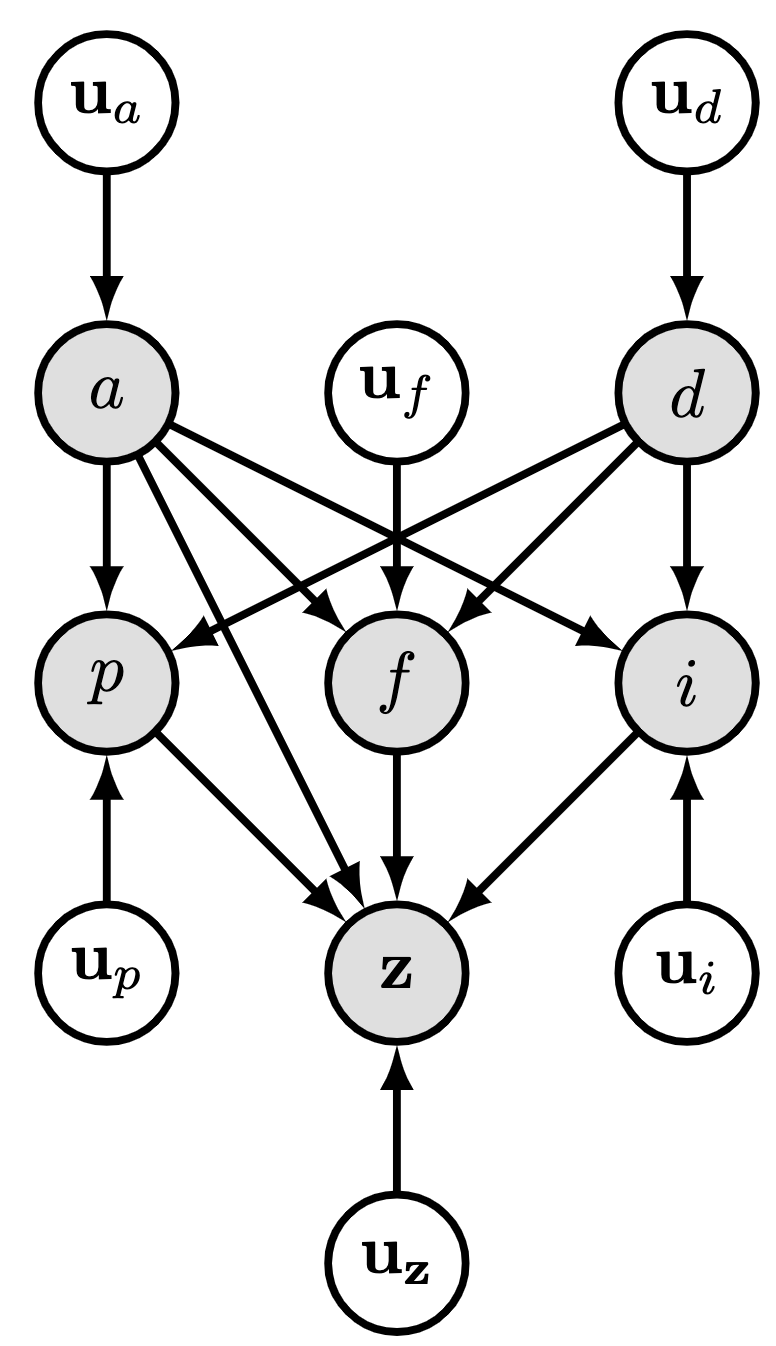}
    \end{subfigure}
\caption{Counterfactual cases. \textbf{Left:} the counterfactual and the differences with original input for each intervention. \textbf{Right:} The assumed causal graph is based on~\cite{sullivan2018role}.  Variables in the graph are age $(a)$, diagnosis $(d)$, Parietal $(p)$, Frontal $(f)$, Insula $(i)$, and latent features of MRI $(\mathbf{z})$. Regarding the difference image, blue refers to negative and light orange to positive differences.}
\label{fig:uncertain}
\end{figure*}

\section{Experiments}\label{sec:experiments}
% 486,150, 626 in UCSF, Lab, and ADNI, respectively
% We will first introduce the data sets and preprocessing steps. Then, the implementation details are described and qualitative and quantitative results are reported. 

% \subsection{Experimental setting}

We train and test our VQ-VAE model on 4583 t1-weighted brain MRIs of 1188 controls from two public datasets: the Alzheimer's Disease Neuroimaging Initiative~\cite{petersen2010alzheimer} (ADNI; 380 subjects, age range: 59 to 91 years, Data Releases: ADNI 1, 2, 3 and GO)  and the National Consortium on Alcohol and Neurodevelopment in Adolescence~\cite{brown2015national} (NCANDA; 808 subjects, age at baseline: 12 to 21 years, maximum age: 27 years, Data Release: NCANDA\_PUBLIC\_6Y\_STRUCTURAL\_V01). Additional details about the two data sets are provided in~\cite{peng2024brainsyn}. We first define the test set by selecting the baseline MRIs of 400 subjects so that they are matched with respect to age and sex, i.e., we separate all samples into 20 groups based on their baseline age and then randomly select 10 females and 10 males for each age group. From the remaining 1919 subjects, we create a training set consisting of 4183 (baseline and follow-up) t1-weighted brain MRIs that is matched with respect to age and sex to the test set. Each MRI is processed using the pipeline described in ~\cite{peng2024brainsyn}, which includes denoising, bias field correction, skull stripping, affine registration to a template (which corrects for difference in head size and thus sex and race), and normalizing intensity values between 0 and 1. The voxel resolution is 1mm and we pad each MRI to spatial resolution $144 \times 176 \times 144$. 

Once training is completed, we freeze the VQ-VAE and train and test the Deep Structural Causal Model (DSCM) on an in-house dataset (PIs: Drs. Pfefferbaum and Sullivan) of 826 T1-weighted MRIs from 421 subjects (age: 25 to 75 years) using five-fold cross-validation. 199 of the 421 subjects are diagnosed with alcohol use disorder (AUD) while the rest are healthy controls. According to  ~\cite{sullivan2018role}, the volumes of the  frontal, insula, and parietal lopes are smaller in those with AUD (compared to controls). Therefore, we build our casual graph, as in Fig.~\ref{fig:uncertain}, based on the five variables age, diagnosis, frontal volume, insula volume, and parietal volume. We evaluate the model with respect to its ability to create counterfactuals and the anatomical plausibility of the synthetic 3D MRIs. 

\subsection{Counterfactual Generation}

Fig.~\ref{fig:uncertain} shows an example of two 3D counterfactuals generated by our approach when applying an intervention \(do()\) to age and diagnosis in our causal graph. According to this figure, our model demonstrates a strong counterfactual generation ability with high fidelity. Intervention on diagnosis reveals a global but subtle changes on the brain. For the age intervention, we can see clear changes in the ventricle, a region known to be significantly influenced by the aging process. 
% For the brain regions, like the frontal lope, the changes are subtle. This may be caused by the limited scale of the dataset.

\begin{figure}[!t]
    \centering
    \includegraphics[width=0.99 \textwidth]{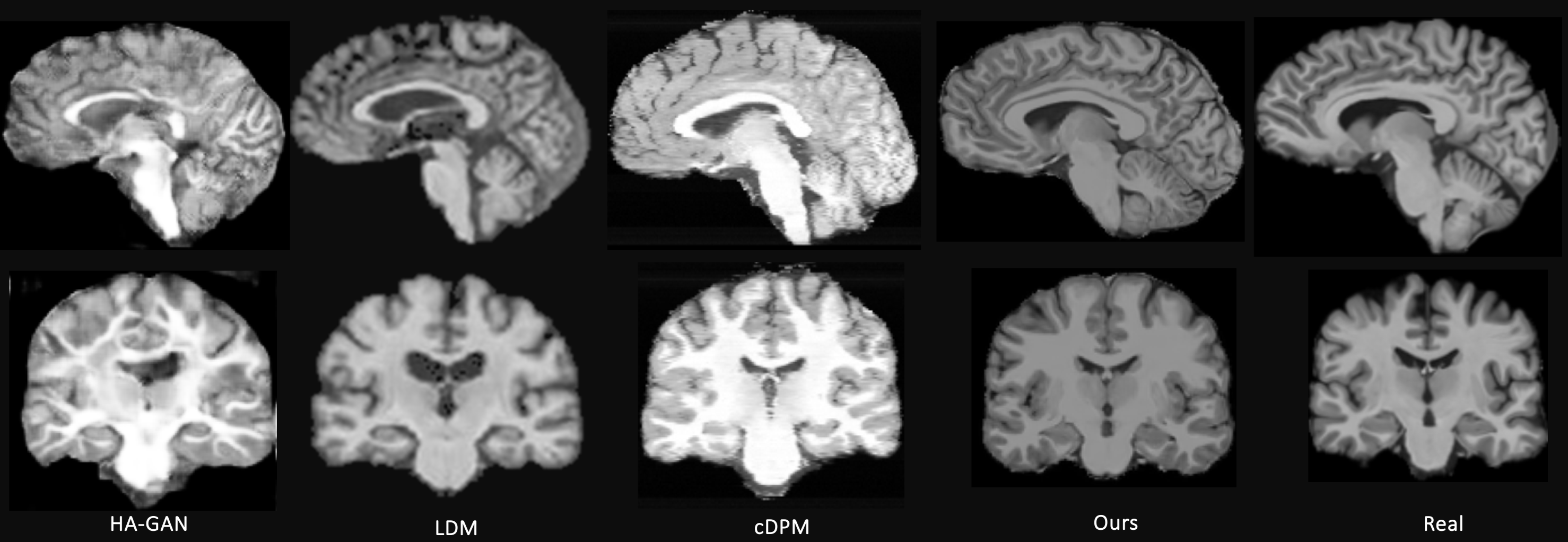}
    \caption{2 views of real MRI \textbf{vs} synthetic MRIs generated by 4 models. Our model can produce visually similar MRI to others but we can do counterfactual generation.}
    \label{fig:showcases}
    \vspace{-0.5cm}
\end{figure}

\subsection{Counterfactual Evaluation}
We created 400 counterfactuals using random interventions. Using the same training setup, we then visually compare one of its generated MRIs to that of the best three baseline approaches according to  ~\cite{peng2024brainsyn}, i.e., HA-GAN~\cite{sun2022hierarchical}) and the diffusion models latent diffusion model (LDM)~\cite{pinaya2023generative} and cDPM~\cite{peng2023generating}). Afterwards, we evaluate its anatomical plausibility as described in ~\cite{peng2024brainsyn}.

According to Fig.~\ref{fig:showcases}, HA-GAN was the only method that did not produce MRIs that looked like those of healthy controls. LDM produces the most noisy MRI and the MRI of cDPM has clear slice artefacts. In contrast, the MRI of our model shows clear gray matter boundaries and looks most similar to the real MRI. Finally, our model is a one-step approach, which is much faster than the diffusion-based method as their is no need for the multi-step diffusion process. 
 
We evaluate the anatomical plausibility of the MRIs produced by our method by comparing  the similarity of their subcortical, cerebellum, and ventricle regions with the real data from ADNI and NCANDA. To this end, we run the Freesurfer pipeline~\cite{fischl2012freesurfer} for both of the 400 real samples and the synthetic ones to obtain the brain parcellations and corresponding volumes of brain regions (recorded in 'aseg' by Freesurfer). We report the effect size of the volume scores of those 24 regions (using Cohen's d coefficient~\cite{cohen2013statistical}, \(d\))  with respect to the real brain MRIs. As shown in Table~\ref{tab:subcortical}, 50\% of the regions have a \(|d|< 0.2\), which means these brain regions are very similar to the real ones. More than 90\% of the 24 brain regions have an effect size that is less than 0.4, which suggests that the observed difference between the real and synthetic is relatively minor or subtle.

\begin{table}[t]
\caption{Subcortical, Cerebellum, and ventricular regions. Cohen's d is reported here as the effect size (\textbf{Effect}). 50\% of the \textbf{ROI}s show strong correlations with the real data as \(|d|< 0.2\).}
\resizebox{\textwidth}{!}{%
\begin{tabular}{rcccccccc}
\hline
  \textbf{ROI}     & \begin{tabular}[c]{@{}c@{}}vessel\\volume\end{tabular} & \begin{tabular}[c]{@{}c@{}}5th\\ventricle\end{tabular}  & \begin{tabular}[c]{@{}c@{}}optic\\chiasm\end{tabular} & \begin{tabular}[c]{@{}c@{}}accumbens\\area\end{tabular} & \begin{tabular}[c]{@{}c@{}}choroid\\plexus\end{tabular}  & \begin{tabular}[c]{@{}c@{}}cc\_mid\\posterior\end{tabular}  & \begin{tabular}[c]{@{}c@{}}cc\\central\end{tabular}  &  \begin{tabular}[c]{@{}c@{}}cc\_mid\\anterior\end{tabular}      \\ \hline
\textbf{Effect} & 0.0065  & -0.0547 & -0.4014  & -0.0028 & 0.2400 & 0.2540 & -0.3216 & -0.1132     \\ \\ 

 \hline
 \textbf{ROI}      & \begin{tabular}[c]{@{}c@{}}cc\\anterior\end{tabular} & \begin{tabular}[c]{@{}c@{}}cc\\posterior \end{tabular} & \begin{tabular}[c]{@{}c@{}}3rd \\ ventricle\end{tabular} & amygdala & \begin{tabular}[c]{@{}c@{}}4th\\ventricle\end{tabular}  & pallidum & \begin{tabular}[c]{@{}c@{}}WM\\hypo.\end{tabular}\  & caudate           \\ \hline
\textbf{Effect}  &   0.3581 & 0.3939 & -0.1804 & -0.1686 & -0.3383 & 0.5588 & -0.0417 & -0.1362 \\\\
\hline
 \textbf{ROI}      & \begin{tabular}[c]{@{}c@{}}hippo\\campus\end{tabular} & putamen & ventraldc & thalamus & \begin{tabular}[c]{@{}c@{}}lateral\\ventricle \end{tabular}   & \begin{tabular}[c]{@{}c@{}}brain\\stem\end{tabular}       & \begin{tabular}[c]{@{}c@{}}cerebellum\\WM\end{tabular}          & \begin{tabular}[c]{@{}c@{}}cerebellum\\cortex\end{tabular}            \\ \hline
\textbf{Effect} & 0.2833 & -0.1373 & 0.6070 & 0.3978 & 0.1306 & 0.2529 & 0.1690 & -0.0748      \\
\hline
\end{tabular}
}
\label{tab:subcortical}
\end{table}

\section{Conclusion}

We propose a novel brain counterfactual for 3D MRI generation, which fully fills three rungs of Pearl’s ladder of causation. The model adeptly performs causal modeling while capable of generating high-quality 3D brain volumes. This is achieved by our latent structural counterfactual model, which contains a deep SCM constructed in a latent space. We then run a GLM on this low dimensional latent space to synthesize counterfactuals, to mitigate significant computational challenges posed by performing interventions directly on the 3D brain MRIs. The generated MRIs show high anatomical resolution, while the counterfactual generation contains an efficient intervention strategy to generate high-quality 3D counterfactuals.

% Future work will involve more data to further improve the cortex-level counterfactual generation. We will also further evaluate its value for neuroscience stuied by te-sting whether the generated counterfactuals can preserve the group-level neuroscience founding or not. 

% A more complex brain causal graph will be constructed to involve more brain regions into the learning process.
%\section*{Acknowledgements}
%\label{sec:acknowledgements}
%\noindent
% \asteriskfill

% \noindent \asteriskfill

 \section{Acknowledgement}
 Collection and distribution of the NCANDA data were supported by NIH funding AA021681, AA021690, AA021691, AA021692, AA021695, AA021696, AA021697. Access to NCANDA data is available via the NIAAA Data Archive, collection 4513 (https://nda.nih.gov/edit\_collection.html?id=4513).
 
 This work was partly supported by funding from the National Institute of Health (MH113406, DA057567, AG084471, AA021697, AA017347, AA010723, AA005965, and AA028840), the DGIST R\&D program of the Ministry of Science and ICT of KOREA (22-KUJoint-02), Stanford School of Medicine Department of Psychiatry and Behavioral Sciences Faculty Development and Leadership Award, the Stanford HAI-Google Research Collaboration, the Stanford HAI Google Cloud Credits, the 2024 Stanford HAI Hoffman-Yee Grant, and the UKRI AI programme, and the Engineering and Physical Sciences Research Council, for CHAI - EPSRC Causality in Healthcare AI Hub (grant no. EP/Y028856/1).

 % This work was partly supported by funding from the National Institute of Health (MH113406, DA057567, AA021697, AA017347, AA010723, AA005965, and AA028840), the DGIST R\&D program of the Ministry of Science and ICT of KOREA (22-KUJoint-02), Stanford School of Medicine Department of Psychiatry and Behavioral Sciences Faculty Development and Leadership Award, and by the Stanford HAI Google Cloud Credit. 

% In addition to the high quality of generated MRI, future works can add condition controls such that generating customized data in a specific surrounding, like brain anatomical variability for different ages. 

% It is maybe problematic to divide the 3D volume into slices. The 3D volume will be considered as independent images. In our model, the attention module is introduced to capture long-range dependency. The results demonstrate that our method can be much better when compared with 3D models. 
%


\bibliographystyle{splncs04}
\bibliography{cites}

\end{document}